\documentclass{article}

\usepackage{microtype}
\usepackage{graphicx}
\usepackage{booktabs} 
\usepackage{listings}
\usepackage{inconsolata}
\usepackage{xcolor}
\usepackage{chngcntr}
\usepackage{apptools}
\usepackage{enumitem}
\usepackage{caption}
\usepackage{subcaption}
\usepackage{multirow}

\definecolor{commentgreen}{rgb}{0,0.6,0} 
\definecolor{keywordred}{rgb}{0.75,0,0}  
\definecolor{stringblue}{rgb}{0,0,0.75}  
\definecolor{black}{rgb}{0,0,0}          

\lstdefinestyle{customstyle}{
    language=Python,
    basicstyle=\ttfamily\scriptsize,  
    keywordstyle=\color{keywordred}\bfseries,  
    stringstyle=\color{stringblue},  
    commentstyle=\color{commentgreen}\itshape,  
    morekeywords={trace}, 
    numbers=none,  
    frame=tb,      
    rulecolor=\color{black}, 
    breaklines=true,
    tabsize=4,
    showstringspaces=false
}

\lstdefinestyle{pythonstyle}{
    language=Python,              
    basicstyle=\ttfamily\footnotesize, 
    keywordstyle=\color{blue},    
    commentstyle=\color{gray},    
    stringstyle=\color{red},      
    numbers=left,                 
    numberstyle=\tiny\color{gray},
    stepnumber=1,                 
    showspaces=false,             
    showstringspaces=false,       
    breaklines=true,              
    breakatwhitespace=true,       
    tabsize=4                     
}

\definecolor{codegreen}{rgb}{0,0.6,0}
\definecolor{codegray}{rgb}{0.5,0.5,0.5}
\definecolor{codepurple}{rgb}{0.58,0,0.82}

\definecolor{backcolour}{RGB}{245,248,250}
\definecolor{emph}{RGB}{166,88,53}
\definecolor{nightblue}{RGB}{9,49,105}
\definecolor{keywords}{RGB}{207,33,46}
\definecolor{lightpurple}{RGB}{130,81,223}

\definecolor{skyblue}{RGB}{86,168,245}
\definecolor{deepblue}{rgb}{0,0,0.5}
\definecolor{deepred}{rgb}{0.6,0,0}
\definecolor{deepgreen}{rgb}{0,0.5,0}

\lstdefinestyle{mystyle}{
    backgroundcolor=\color{backcolour},    %
    commentstyle=\color{codegreen},
    keywordstyle=\color{keywords},
    stringstyle=\color{nightblue},
    basicstyle=\fontsize{7}{8}\ttfamily,
    breakatwhitespace=true,         
    breaklines=true,                 
    captionpos=b,                    
    keepspaces=true,                 
    numberstyle=\tiny\color{codegray},
    numbersep=2pt,                  
    showspaces=false,                
    showstringspaces=false,
    showtabs=false,                  
    tabsize=2,
    emph={bundle,@trace,trace,backward, instruction, code, documentation, variables, constraints,inputs,others,outputs,feedback,actual_problem_instance,variable1_name,variable2_name,variable1_value1,variable1_value2,variable2_value1,variable2_value2,feedback_1,feedback_2,node,@trace_class},
    emphstyle={\color{codepurple}},
    linewidth=1\columnwidth,
    frame=tb,    
    xrightmargin=0pt,
    xleftmargin=0.23cm,
    numbers=left,
    aboveskip=0.2cm,
    belowskip=0.1cm,
}

\usepackage{hyperref}

\usepackage[accepted]{icml2025_pral}

\usepackage{amsmath}
\usepackage{amssymb}
\usepackage{mathtools}
\usepackage{amsthm}

\usepackage[capitalize,noabbrev]{cleveref}

\theoremstyle{plain}

\theoremstyle{definition}

\theoremstyle{remark}

\usepackage[textsize=tiny]{todonotes}

\icmltitlerunning{Learning Game-Playing Agents with Generative Code Optimization}

\begin{document}

\twocolumn[
\icmltitle{Learning Game-Playing Agents with Generative Code Optimization}



\icmlsetsymbol{equal}{*}

\begin{icmlauthorlist}
\icmlauthor{Zhiyi Kuang}{stanford}
\icmlauthor{Ryan Rong*}{stanford}
\icmlauthor{YuCheng Yuan*}{stanford}
\icmlauthor{Allen Nie}{stanford}
\end{icmlauthorlist}

\icmlaffiliation{stanford}{Department of Computer Science, Stanford University, Stanford, CA, USA}

\icmlcorrespondingauthor{Zhiyi Kuang}{kuangzy@stanford.edu}

\icmlkeywords{Agent Learning, Large Language Model Optimizer}

\vskip 0.3in
]



\printAffiliationsAndNotice{\icmlEqualContribution} 

\begin{abstract}
We present a generative optimization approach for learning game-playing agents, where policies are represented as Python programs and refined using large language models (LLMs). Our method treats decision-making policies as self-evolving code, with current observation as input and an in-game action as output, enabling agents to self-improve through execution traces and natural language feedback with minimal human intervention. Applied to Atari games, our game-playing Python program achieves performance competitive with deep reinforcement learning (RL) baselines while using significantly less training time and much fewer environment interactions. This work highlights the promise of programmatic policy representations for building efficient, adaptable agents capable of complex, long-horizon reasoning.
\end{abstract}

\section{Introduction}
A core challenge in AI is developing agents that learn complex tasks efficiently and in ways that are interpretable to humans. While traditional reinforcement learning (RL) has achieved impressive results across domains including video games \cite{mnih2013playing, schulman2017proximal}, robotics \cite{xu2023unidexgrasp}, embodied intelligence \cite{gupta2021embodied}, and autonomous vehicles \cite{kiran2021deep}, these methods often require millions of interactions and produce opaque policies that are hard to verify--especially problematic in safety-critical applications. Atari games, for example, remain a longstanding benchmark where standard algorithms like PPO \cite{schulman2017proximal} demand heavy sampling to perform well. These challenges have spurred growing interest in alternate approaches that improve sample efficiency and transparency.

Programmatic policies--explicit code that defines agent behavior--offer interpretability, modularity, and formal verifiability. If such policies could be optimized as efficiently as neural networks, they would enable agents whose decisions can be inspected, tested, and reused--supporting safety and generlization across tasks. Demonstrating this in Atari would provide strong evidence that code is a viable representation for sequential decision-making.

However, optimizing programs is fundamentally different from tuning neural weights. Code is non-differentiable, making gradient-based methods inapplicable, while brute-force search \cite{abdollahi2023can} or evolutionary methods \cite{cui2021alphaevolve} struggle with combinatorial search. Conventional RL methods \cite{mnih2013playing, schulman2017proximal} provide limited feedback on which parts of the agent logic caused failure. Naïve LLM prompting often yields brittle, one-shot scripts with poor execution grounding. 

While recent work has explored using LLMs for code generation and optimization in various domains \cite{skreta2023errors, xia2023automated, huang2024effilearner, ishida2024langprop}, their application to agent policy optimization--guided by structural execution feedback--remains underexplored. Our approach departs from prior work by: (i) treating the policy code itself as the object of optimization, (ii) extracting rich execution traces to localize failure points, and (iii) prompting an LLM iteratively with graph-based backtracing to propose meaningful code updates.

Our approach treats Atari gameplay as a programmatic control problem, where policies are written as modular Python programs and refined through LLM-guided updates. Using the Trace framework \cite{cheng2024trace}, we execute policy rollouts in the environment and optimize policies based on structured feedback from gameplay outcomes. A key difference from prior work is the complexity of our domain: while Trace focuses on short-horizon tasks (e.g. Meta-World environments with at most 10 decision steps), Atari games require hundreds and thousands of steps per episode with sparse rewards, introducing longer temporal dependencies and credit assignment challenges. Despite this, our method enables interpretable and efficient learning, with agents reaching competitive performance while remaining human-readable by design. Our main contributions can be summarized as follows: 
\vspace{-4mm}
\begin{itemize}[leftmargin=1.5em]
    \item We present the first application of Python code-based policy optimization on Atari games using generative LLM updates.
    \item We demonstrate that this method can learn long-horizon, sparse-reward policies through natural language feedback and computational graph trace reasoning, achieving competitive performances with deep RL baselines while using significantly less training time (-98\% to -52\%) and fewer environment interactions.
\end{itemize}

\section{Related Work}

\paragraph{Reinforcement Learning for Atari Games.}
Model-free RL algorithms dominate Atari gameplaying benchmarks, such as DQN \cite{mnih2013playing}, PPO \cite{schulman2017proximal}, and distributional variants such as IQN \cite{dabney2018implicit} and C51 \cite{bellemare2017distributional}. To curb sample inefficiency, model-based agents combine world models with planning, including SimPLe \cite{kaiser2019model}, DreamerV2 \cite{hafner2020mastering}, and MuZero \cite{schrittwieser2020mastering}. Transformer-based approaches \cite{chen2021decision,lee2022multi} and object-centric approach \cite{delfosse2024ocatari} have also been explored. We use generative optimization to overcome high sample complexity common in traditional methods. 

\paragraph{LLMs for Gameplaying.}
The application of LLM to gameplaying has gained significant attention recently. LLM exhibits solid ability to understand logic across a wide variety of games, such as Slay the Spire \cite{bateni2024language}, Minecraft \cite{wang2023voyager}, Pokémon \cite{karten2025pok}, StarCraft II \cite{ma2024large}, NetHack \cite{jeurissen2024playing}, simplified Maze \cite{sanchez2024controlling}, Rock Paper Scissors \cite{vidler2025playinggameslargelanguage}, and Negotiation games \cite{hua2024gametheoreticllmagentworkflow}. While previous approaches focus on showcasing LLM's ability to zero-shot these games, our study examines LLM's capacity of iteratively refining game policy based on fine-grained environmental feedback.

\paragraph{Methods of Code and Policy Optimization.}
Prior pre-LLM works have attempted to improve policy generalization and verifiability by synthesizing structured policies, such as LEAPS \cite{trivedi2021learning}, VIPER \cite{bastani2018verifiable}, and learning programmatic state
machine policies \cite{inala2020synthesizing}.
Building on this, general-purpose LLMs have demonstrated substantial reasoning ability for code control flow \cite{chen2021evaluating, ouyang2022training,  roziere2023code}. Furthermore, various studies substantiated LLM's potential for code-optimization, such as LLM-based iterative optimizers including LangProp \cite{ishida2024langprop}, EffiLearner \cite{huang2024effilearner}, CLARIFY \cite{skreta2023errors}, and AutoPatch \cite{acharya2025optimizingcoderuntimeperformance}. LLM-based generative optimization is able to iteratively refine solutions based on various forms of feedback, yet its potential for optimizing agent policy for gaming remains underexplored. Our work is the first to use generative optimization to improve agent policy performance in Atari games.

\section{Approach}
We propose an LLM-based generative optimization approach for developing Atari game-playing agents, where policies are represented as modular Python programs and optimized within the Trace \cite{cheng2024trace} framework. Unlike traditional RL algorithms that train neural network-based policies, our method treats policy components such as action selection as trainable functions written in code. Trace captures detailed records of the agent's interactions with the environment (called "execution traces"), allowing a large language model to iteratively refine the policy based on this structured interaction data and natural language feedback. This allows interpretable and efficient policy learning in complex, long-horizon environments.
\vspace{-3mm}
\begin{figure}[htbp]
\begin{subfigure}[t]{0.48\columnwidth}
\begin{lstlisting}[language=Python,
breaklines=true,
showstringspaces=false,
basicstyle=\fontsize{5.7}{6.3}\selectfont\ttfamily,
numbers=none,
escapechar=!]
class Policy(trace.Module):
 def __call__(self, obs):
   pred = self.predict_pos(obs)
   action = self.act(pred, obs)
   return action
    
 @trace.bundle(trainable=True)
 def predict_pos(self, obs):
   """
   Estimate ball trajectory from observation
   """
    
 @trace.bundle(trainable=True)
 def act(self, pred, obs):
   """
   Move paddle towards prediction
   """
\end{lstlisting}
\caption*{\textbf{(a)} Trainable policy with prediction and action functions.}
\label{fig:policy-class}
\end{subfigure}
\hfill
\begin{subfigure}[t]{0.5\columnwidth}
\begin{lstlisting}[language=Python,
breaklines=true,
showstringspaces=false,
basicstyle=\fontsize{6}{6.3}\selectfont\ttfamily,
numbers=none,
escapechar=!]
policy = Policy()
params = policy.parameters()
optimizer = trace.Optimizer(params)
env = TracedEnv()

for i in range(iters):
  # Forward pass
  traj = rollout(env, policy)
  perf = evaluate(env, policy)
  feedback = compute_feedback(perf)
  target = traj['obs'][-1]

  # Backward pass
  optimizer.zero_feedback()
  optimizer.backward(target, feedback)
  optimizer.step()
\end{lstlisting}
\caption*{\textbf{(b)} Optimization loop using Trace.}
\label{fig:opt-loop}
\end{subfigure}
\caption{\textbf{Policy Learning with Trace.} The agent's behavior is defined by (a) trainable, modular functions, and (b) refined through rollout-based optimization using structured feedback.}
\end{figure}


\subsection{Object-Centric Atari Representations}
\begin{figure}
    \centering
    \includegraphics[width=1\linewidth]{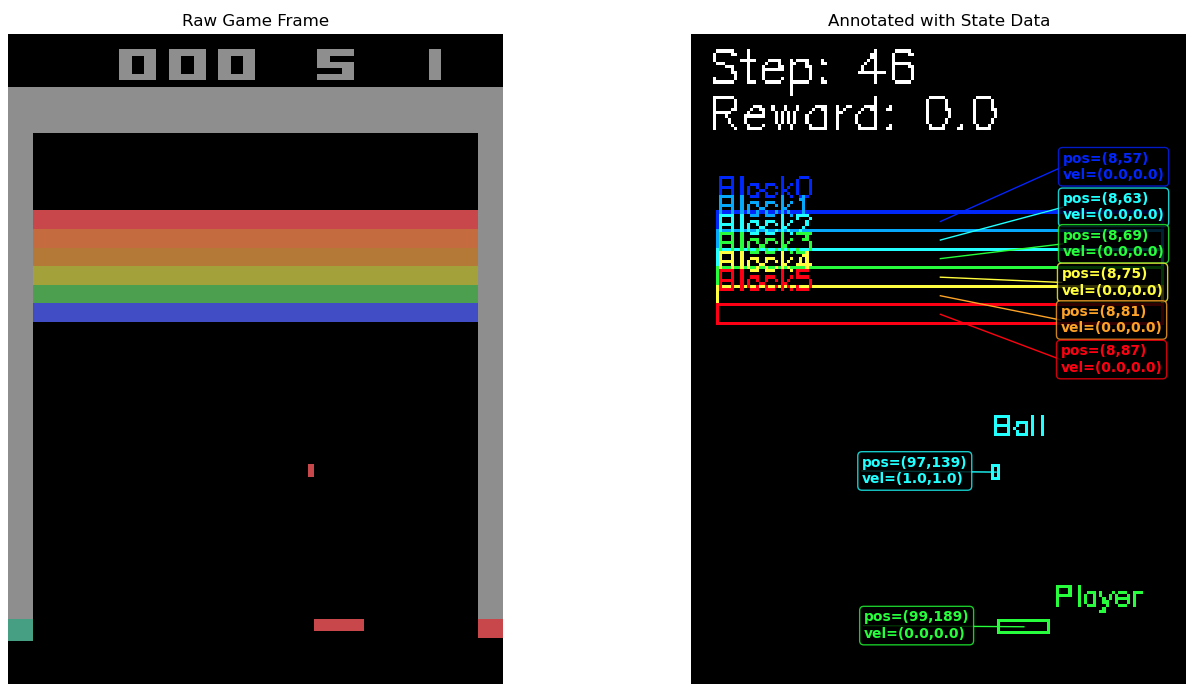}
    \caption{\textbf{Visual Comparison of the Original Atari Game Screen (Left) and Object-Centric Representation (Right) in Breakout.} The object-centric view provides a compact and interpretable state abstraction.
    This representation allows our agents to reason over gameplay dynamics efficiently.}
    \label{fig:ocatari}
\end{figure}

We use Object-Centric Atari environments (OCAtari) \cite{delfosse2024ocatari} to convert pixel-based observation from Arcade Learning Environment (ALE) \cite{bellemare13arcade} to object-level representations. OCAtari extracts key information for each game object, including coordinates ($x, y$), size (width, height), and velocity ($dx, dy$), rewards, and game termination status (e.g., ``lives") (see Figure \ref{fig:ocatari}). We generate this data on-the-fly during training and do not apply additional transformations.

\subsection{Generative Optimization}
\paragraph{Agent Design.} We represent the agent's policy as a modular Python program, structured around a high-level \textbf{plan-act} interface. Each policy consists of decision-making functions--such as planning trajectories or selecting actions--which are annotated with \lstinline[style=pythonstyle]{@trace.bundle(trainable=True)} to mark them as optimizable by the Trace framework. While the core interface remains consistent across games, for each game, we instantiate custom planning and acting components to reflect specific mechanics and decision-making process.

For both Pong and Breakout, the policy includes a \lstinline[style=pythonstyle]{predict_ball_trajectory} function that estimates the ball's future position. This prediction informs a \lstinline[style=pythonstyle]{select_action} function that determines how the paddle should move. For Breakout, we introduce an additional \lstinline[style=pythonstyle]{generate_paddle_target} component to prioritize targeting high-value bricks and forming tunnel strategies, adding a layer of strategic planning as a heuristic to guide the generative optimization. In Space Invaders, the policy is decomposed into \lstinline[style=pythonstyle]{decide_shoot} and \lstinline[style=pythonstyle]{decide_movement} functions, allowing the agent to independently control when to fire and how to move the player avatar.

\paragraph{Learning Design.} 
We train agents in an episodic reinforcement learning setup, where each iteration consists of a single rollout. During a rollout, the agent observes the environment, selects actions, and receives rewards. This rollout trajectory is traced end-to-end and is provided as execution traces to an LLM-based optimizer along with natural language feedback derived from a full-length evaluation episode (e.g. $\sim$4000 steps). 

We use OptoPrime \cite{cheng2024trace}, a generative optimizer that updates the agent's policy by modifying its trainable code components. The number of steps per rollout is limited by the LLM's context window, which must accommodate the trajectory, observations, and function definitions. To fit within this token budget, we cap rollouts at 400 steps for Pong, 300 for Breakout, and 15 for Space Invaders.

Our design requires minimal human intervention: the user defines the high-level function interfaces, writes docstrings and starter code, and configures automatic feedback. All policy improvements thereafter are generated autonomously by the LLM through iterative optimizations. 

\paragraph{Staged Feedback Design.} We observe that using only reward-based feedback from training rollouts often leads to performance plateaus--especially in games with evolving dynamics. For instance, in Breakout, bricks in the upper rows deflect the ball at higher speeds, creating a distribution shift between the training context (primarily lower bricks) and the evaluation context (including higher bricks). This observation inspires two feedback design choices: 1) we evaluate the performance of the agent with longer evaluation rollouts and use that reward as feedback to the generative optimizer; 2) we provide \textit{staged feedback} to instruct the model to pay attention to different game mechanisms or share high-level winning strategies. To implement staged feedback, we design natural language responses for different levels of agent performance (Table \ref{tab:feedback-structure}).
\vspace{-5mm}
\begin{table}[htbp]
\caption{\textbf{Staged Feedback for the Pong Agent at Different Performance Levels.}}
\label{tab:feedback-structure}
\vskip 0.15in
\renewcommand{\arraystretch}{1.3}
\resizebox{\columnwidth}{!}{
\begin{tabular}{p{3cm}p{10cm}}
\toprule
\textbf{Performance Level} & \textbf{Feedback} \\
\midrule
High \newline(Reward $\geq$ 19) & "Good job! You're close to winning the game! You're scoring 20 points against the opponent, only 1 points short of winning." \\
\midrule
Medium \newline(0 $<$ Reward $<$ 19) & "Keep it up! You're scoring 12 points against the opponent but you are still 9 points from winning the game. Try improving paddle positioning to prevent opponent scoring." \\
\midrule
Low \newline (Reward $\leq$ 0) & "Your score is $-5$ points. Try to improve paddle positioning to prevent opponent scoring." \\
\bottomrule
\end{tabular}
}
\end{table}

\section{Evaluation}

\subsection{Results}

We evaluate our approach in three classic Atari environments: Pong, Breakout, Space Invaders. The training configuration is reported in Table \ref{tab:training-configs} and the environment setup is reported in Table \ref{tab:environment-configs}. We compare the performance of our approach with open-source implementations of deep RL baselines, including DQN \cite{mnih2013playing} and PPO \cite{schulman2017proximal}, as well as human-level performance benchmarks. We demonstrate that our approach can match some existing deep RL baselines while requiring significantly less training time and fewer environment interactions (Table \ref{tab:atari-main-table}). 

\begin{table}[htbp]
\caption{\textbf{Comparison of Atari Performance and Training Time.} Due to high variations in reported results across papers, we compare against standardized baselines from open-source RL implementations \cite{huang2022cleanrl, huang202237}. RL algorithms are trained with 8 parallel environment instances, while our agent uses only 1. For reference, highly optimized deep RL with 32 environment instances can reach a Breakout score of $\sim$450 in 33m, see Appendix~\ref{sec:app:deep-rl-result}. }
\label{tab:atari-main-table}
\vskip 0.15in
\centering
\resizebox{\columnwidth}{!}{
\begin{tabular}{lcccccc}
\toprule 
Game & Learned Agent & DQN (Time) & PPO (Time) & Human \\
\midrule
Pong           & \textbf{21} (43m)  & 20 (10h 6m)    & 19 (2h 24m)    & 14.59 \\
Breakout       & \textbf{353} (1h 31m) & 302 (26h 54m)   & 443 (3h 8m)   & 30.47  \\
Space Invaders & \textbf{1200} (36m)         & 1383 (26h 52m) & 939 (5h 39m)     & 1668.67 \\
\bottomrule
\end{tabular}
}
\end{table}

\begin{table}[htbp]
\caption{\textbf{Atari Game-Specific Experiment Configurations.}}
\label{tab:environment-configs}
\vskip 0.15in
\centering
\resizebox{\columnwidth}{!}{
\begin{tabular}{l|ccc}
\toprule
\textbf{Parameter} & \textbf{Breakout} & \textbf{Pong} & \textbf{Space Invaders} \\
\midrule
Rollout horizon & 300 steps & 400 steps & 15 steps \\
Action space & LEFT/RIGHT/NOOP & UP/DOWN/NOOP & LEFT/RIGHT/FIRE/NOOP \\
Env special mechanics & Auto-fire on life loss & None & Fire cooldown \\
\bottomrule
\end{tabular}
}
\end{table}

\subsection{Emergent Gameplay Understanding}
OptoPrime \cite{cheng2024trace} shows a surprising ability to infer underlying game dynamics and constraints from sparse trajectory data. While we provide high-level guidance through docstrings, we experiment with deliberately omitting specific implementation details such as boundary positions and collision physics. OptoPrime is able to correctly recover the missing information by analyzing traced trajectory data. For example, Figure \ref{fig:llm-infered-code} illustrates how OptoPrime identifies the exact position of the right wall ($x=152$) by observing ball position and velocity changes across multiple steps. It also learns accurate ball physics such as bounce mechanics without explicitly being told these details. This highlights LLM's ability for causal reasoning over long, sparse sequence.

\subsection{Code Complexity Analysis}
To analyze how agent evolves, we track code complexity over optimization steps. As shown in Table \ref{tab:code-analysis}, the policies grow significantly in length and structual complexity across iterations, measured by lines of code (LOC), cyclomatic complexity (Comp.), and the maximum nested \texttt{if} depth (N. Ifs). Cyclomatic complexity \cite{mccabe76} quantifies the number of independent execution paths through the code. Final policies are consistently more complex than initial scaffold, reflecting progressive refinement. Notably, complexity often plateaus or slightly decreases in later iterations, suggesting the model reorganizes logic for efficiency rather than continues to expand it indefinitely. 

\begin{table}[htbp]
\caption{\textbf{Code Metrics for Selected Policy Stages.} ``It." denotes the iteration number corresponding to the policy stage.}
\label{tab:code-analysis}
\vskip 0.15in
\centering
\small 
\setlength{\tabcolsep}{4pt} 
\begin{tabular}{@{}llrrr@{}} 
\toprule
Game & Policy Stage & LOC & Comp. & N. Ifs \\
\midrule
\multirow{3}{*}{Space Invaders} & Initial (It. 0) & 117 & 20 & 3 \\
                               & Intermediate (It. 9) & 146 & 28 & 3 \\
                               & Best (It. 10) & 146 & 28 & 3 \\
\midrule
\multirow{3}{*}{Pong} & Initial (It. 0) & 49 & 2 & 1 \\
                      & Intermediate (It. 6) & 94 & 9 & 1 \\
                      & Best (It. 11) & 131 & 16 & 2 \\
\midrule
\multirow{3}{*}{Breakout} & Initial (It. 0) & 95 & 5 & 1 \\
                          & Intermediate (It. 11) & 134 & 24 & 3 \\
                          & Best (It. 20) & 125 & 24 & 3 \\
\bottomrule
\end{tabular}

\end{table}

\subsection{Ablation Study of Staged Feedback}
To evaluate the impact of the staged feedback design and full-game evaluation, we perform an ablation study on the game \textbf{Pong}, comparing two conditions: (1) using only reward-based feedback from short training rollouts, (2) incorporating full-game evaluation reward as feedback. Since training rollouts are short (e.g. 400 steps) and limited by the context window of the LLM, they often capture only fragments of a full game. Pong is a relatively simple environment compared to Breakout or Space Invaders. However, as Table \ref{tab:pong-feedback-comparison} shows, feedback derived solely from traced rollouts leads to performance plateaus, even in this simpler setting.
\begin{table}[!htbp]
\caption{\textbf{Impact of Full-Game Staged Feedback on Performance in Pong.} Despite Pong's relatively simplicity, using only traced rollout reward feedback leads to performance plateaus. This demonstrates the importance of providing long-horizon feedback in overcoming context limitations of the LLM for successful policy optimization.}
\label{tab:pong-feedback-comparison}
\vskip 0.15in
\centering
\resizebox{\columnwidth}{!}{
\begin{tabular}{lcc}
\toprule
\textbf{Feedback Type} & \textbf{Max Performance}  \\
\midrule
Rollout-Only Feedback & 7\\
Rollout + Full-Game Staged Feedback & \textbf{21} \\
\bottomrule
\end{tabular}
}

\end{table}

\section{Discussions and Limitations}
We demonstrate that generative code optimization can produce game-playing agents that achieve performance competitive with deep RL methods using significantly less training time and environmental interaction. By refining modular Python policies through execution traces and structured feedback, our approach demonstrates interpretable and sample-efficient learning in long-horizon, sparse-reward tasks. 

However, our approach has limitations: LLMs can introduce occasional unstable edits and the performance depends on carefully crafted prompts due to the context window constraint of current models. Nonetheless, this work introduces a novel framework for agent learning that combines programmatic reasoning and language-based optimization in sparse-reward setting.

\section*{Impact Statement}

This paper presents work whose goal is to advance the field
of Machine Learning. There are many potential societal
consequences of our work, none which we feel must be
specifically highlighted here.

\nocite{langley00}

\bibliography{references}
\bibliographystyle{icml2025}

\newpage
\appendix
\renewcommand{\thefigure}{A.\arabic{figure}}
\renewcommand{\thetable}{A.\arabic{table}}
\setcounter{figure}{0}
\setcounter{table}{0}
\onecolumn
\section{Atari Game Setup}

\paragraph{Pong} In Pong, the player controls a paddle on the right side of the screen to deflect the ball into the enemy's goal. The player scores a point if the enemy misses the ball. The game ends when one side scores 21 points.

\paragraph{Breakout} In Breakout, the player moves a bottom paddle horizontally to deflect a ball that scores against brick walls upon contact. The brick wall consists of six rows of different colored bricks, with higher bricks worth more points. Hitting higher bricks would deflect the ball faster, increasing the difficulty in catching the ball. The player wins after scoring 864 points. The player loses one life when failing to catch the ball and the ball moves out of range. The player has five lives in total.

\paragraph{Space Invaders} In Space Invaders, the player controls a turret to shoot down aliens and alien ship that float around the screen, while dodging the aliens' attacks. There are three shields that can absorb both the player and the aliens' attacks. There can only be 1 player bullet on the field at a time, and the player has three lives. 

The training configuration is reported in Table \ref{tab:training-configs}.
\begin{table}[htbp]
\caption{\textbf{Environment and Training Configurations.}}
\label{tab:training-configs}
\vskip 0.15in
\centering
\begin{tabular}{lc}
\toprule
\textbf{Parameter} & \textbf{Value} \\
\midrule
Environment Name & \{env\}-*NoFrameskip-v4 \\
Action Repeat (Frameskip) & 4 \\
Sticky Action Probability & 0.0 \\
\midrule
Optimization Iterations & 20 \\
Rollout Length & 15/300/400 steps \\
Memory Size (Optimizer Context) & 5 \\
Evaluation Episode Length & $\sim$4000 steps \\
LLM Optimizer & OptoPrime  \\
LLM Backend & Claude-3-5-sonnet-20241022-v2:0 \\
Access Date & Feb-May 2025 \\
\bottomrule
\end{tabular}

\end{table}

\begin{figure}
    \centering
    \includegraphics[width=0.8\linewidth]{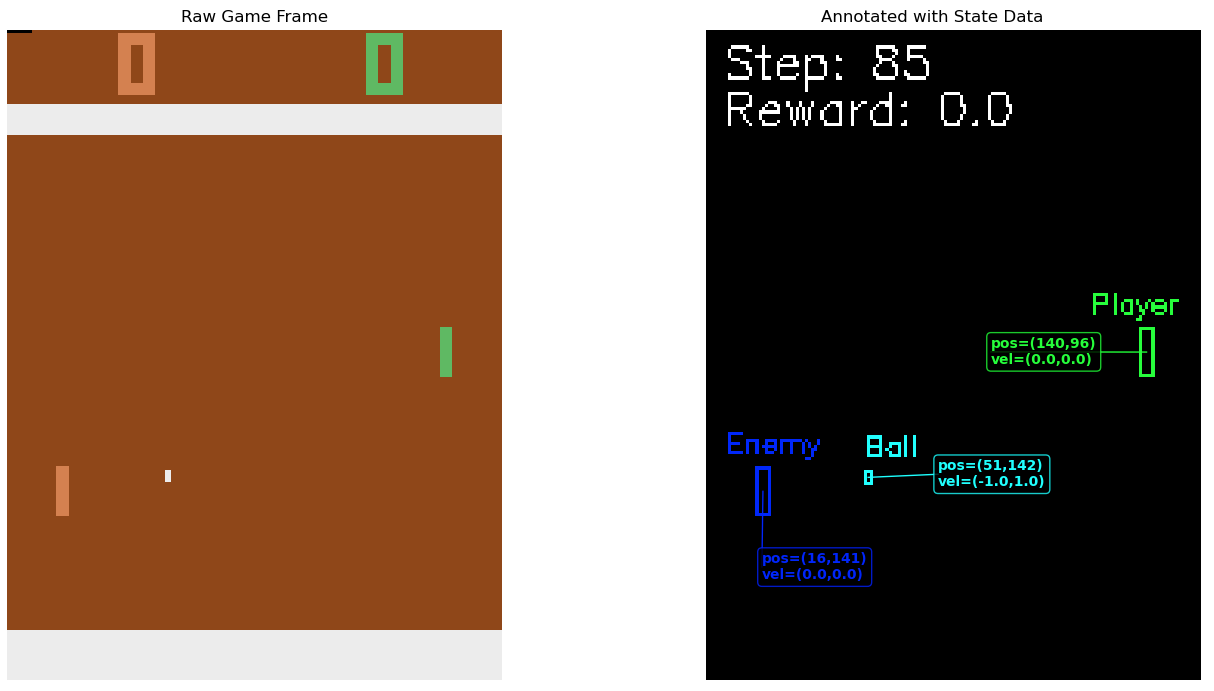}
    \caption{\textbf{Visualization of Pong Atari Game Screen (Left) and Object-Centric Representation (Right).}}
    \label{fig:pong_visualization}

    \vspace{1.5em}
    
    \includegraphics[width=0.8\linewidth]{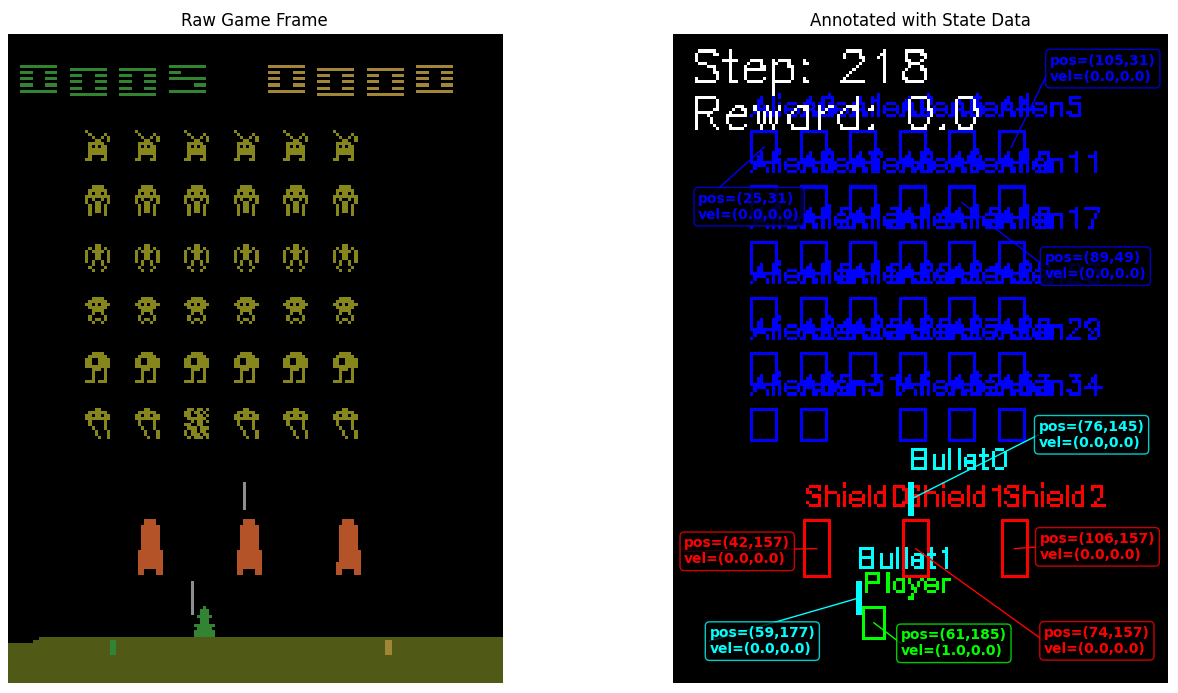}
    \caption{\textbf{Visualization of Space Invaders Atari Game Screen (Left) and Object-Centric Representation (Right).}}
    \label{fig:space_invaders_visualization}
\end{figure}

\section{Agent Design Details}

\begin{figure}[h]
\centering
\begin{subfigure}[t]{\textwidth}
    \centering
    \includegraphics[width=0.8\linewidth]{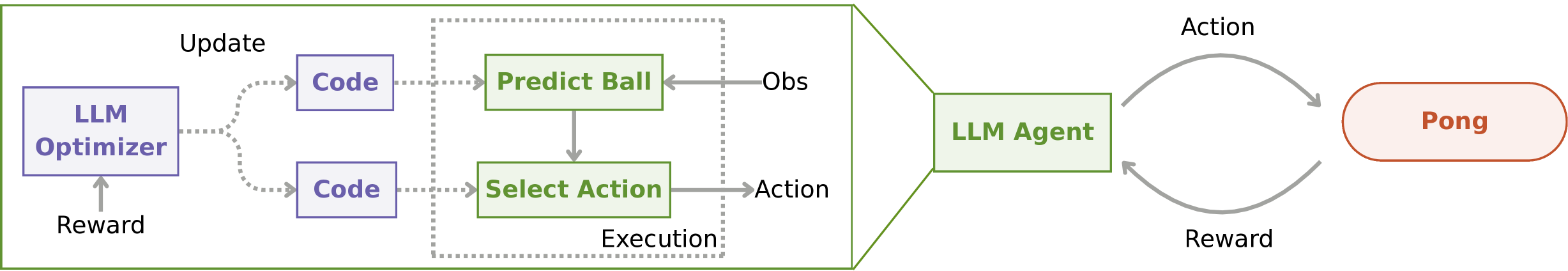}
    \caption{Pong Agent}
    \label{fig:atari_pong_agent_graph}
\end{subfigure}%
\hfill
\begin{subfigure}[t]{\textwidth}
    \vspace{2mm}
    \centering
    \includegraphics[width=0.8\linewidth]{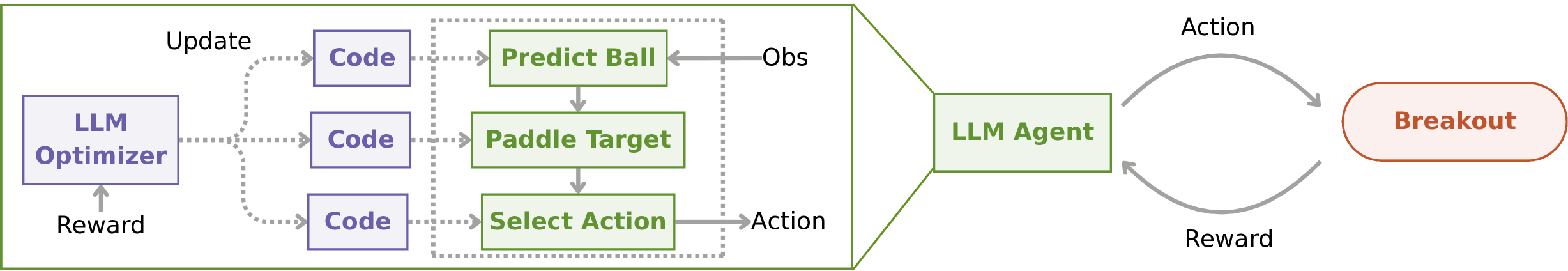}
    \caption{Breakout Agent}
    \label{fig:atari_breakout_agent_graph}
\end{subfigure}
\hfill
\begin{subfigure}[t]{\textwidth}
    \vspace{2mm}
    \centering
    \includegraphics[width=0.8\linewidth]{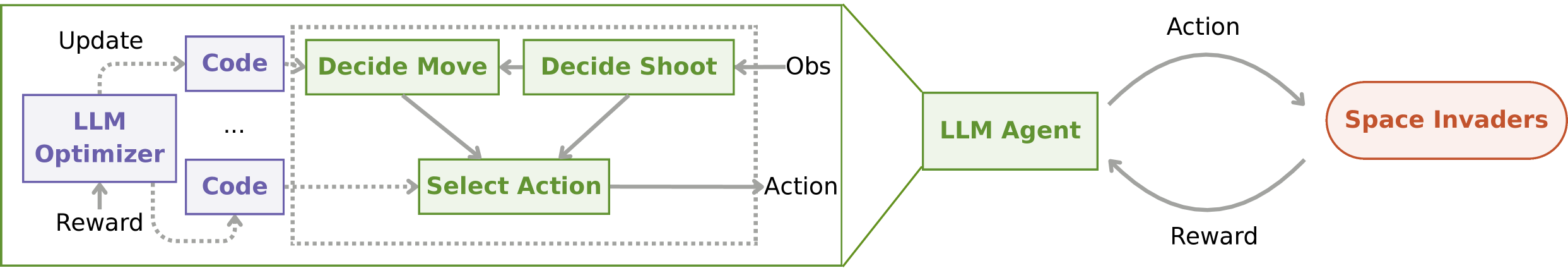}
    \caption{Space Invaders Agent}
    \label{fig:atari_space_invaders_agent_graph}
\end{subfigure}
\caption{\textbf{Agent Design to Play Atari Games}. }
\label{fig:atari_agent_graphs}
\end{figure}

\ref{fig:atari_agent_graphs} is a graphical visualization of the high-level design of the Pong, Breakout, and Space Invaders agents. 

\section{Feedback Design Details}
We provide game-specific feedback instructions when the agent reaches different reward regions. Although the maximum achievable score in games like Breakout (864) and Space Invader (typically several thousands) is significantly higher, we deliberately define the ``High" performance threshold at a lower reward level (e.g. $\geq 300$ for Breakout). This choice reflects the relatively short training horizon during each optimization iteration (15/300/400 steps), which is constrained by the context window size of the LLM. Setting a lower threshold allows the feedback to remain meaningful and actionable within the context of short-term learning progress, while still guiding the agent toward longer-term strategies over multiple iterations. Staged feedback for the Breakout agent and Space Invaders agent are shown in Table \ref{tab:feedback-breakout}, \ref{tab:feedback-space}. 

\begin{table}[htbp]
\caption{\textbf{Staged Feedback for the Breakout Agent at Different Performance Levels.}}
\label{tab:feedback-breakout}
\vskip 0.15in
\centering
\renewcommand{\arraystretch}{1.3}
\begin{tabular}{p{5cm}p{5cm}}
\toprule
\textbf{Performance Level} & \textbf{Feedback} \\
\midrule
High \newline(Reward $\geq$ 300) & "Good job! You're close to winning the game! You're scoring 320 points against the opponent, try ensuring you return the ball, only 30 points short of winning." \\
\midrule
Medium \newline(0 $<$ Reward $<$ 300) & "Keep it up! You're scoring 50 points against the opponent but you are still 300 points from winning the game. Try improving paddle positioning to return the ball and avoid losing lives." \\
\midrule
Low \newline (Reward $\leq$ 0) & "Your score is 0 points. Try to improve paddle positioning to return the ball and avoid losing lives." \\
\bottomrule
\end{tabular}

\end{table}

\begin{table}[htbp]
\caption{\textbf{Staged Feedback for the Space Invaders Agent at Different Performance Levels.}}
\label{tab:feedback-space}
\vskip 0.15in
\centering
\renewcommand{\arraystretch}{1.3}
\begin{tabular}{p{5cm}p{5cm}}
\toprule
\textbf{Performance Level} & \textbf{Example Feedback} \\
\midrule
High \newline(Reward $\geq$ 1000) & "Great job! You're performing well with an average score of 1005. Try to score more even more points" \\
\midrule
Medium \newline(500 $<$ Reward $<$ 1000) & "Good progress! Your average score is 570. Focus on better timing for shooting and avoiding enemy projectiles." \\
\midrule
Low \newline (Reward $\leq$ 500) & "Your average score is 270. Try to improve your strategy for shooting aliens and dodging projectiles." \\
\bottomrule
\end{tabular}

\end{table}

\section{Emergent Gameplay Understanding of LLM Optimizer}
Figure \ref{fig:llm-infered-code} illustrates that LLM Optimizer can infer underlying game dynamics and constraints from sparse trajectory data, without being explicitly told these details.
\begin{figure}[htbp]
\centering
\begin{minipage}[t]{0.52\textwidth}
\begin{subfigure}[t]{\textwidth}
\begin{lstlisting}[language=Python,
breaklines=true,
showstringspaces=false,
basicstyle=\fontsize{8}{9}\selectfont\ttfamily,
numbers=none,
escapechar=!]
class BreakoutPolicy(tace.Module):
  def predict_ball_trajectory(self, obs):
    """
    Game setup:
    - Screen dimensions:
      - Left wall: x=9
      - Right wall:
    [Additional docstring sections omitted]
    """
    # Code omitted 

\end{lstlisting}
\caption*{\textbf{(a)} The user provides a partially specified docstring in the policy; the right wall is unspecified.}
\end{subfigure}
\begin{subfigure}[t]{\textwidth}
\begin{lstlisting}[language=Python,
breaklines=true,
showstringspaces=false,
basicstyle=\fontsize{8}{9}\selectfont\ttfamily,
numbers=none,
escapechar=!]
class BreakoutPolicy(trace.Module):
  def predict_ball_trajectory(self, obs):
    """
    Game setup:
    - Screen dimensions:
      - Left wall: x=9
      - Right wall: x=152
    [Additional docstring sections omitted]
    """
    # Code omitted
\end{lstlisting}
\caption*{\textbf{(b)} OptoPrime infers the missing right wall location ($x=152$) from the observed ball trajectory, and updates the docstring accordingly. Additional function logic (not shown) is also completed to implement calculations based on bouncing logic.}
\end{subfigure}
\end{minipage}
\hfill
\begin{minipage}[c]{0.45\textwidth}
\vspace{24em}
\centering
\renewcommand{\arraystretch}{1.2}
\centering
\begin{subfigure}[c]{\textwidth}
\begin{tabular}{c|cc}
        Step & Ball $x$ & Ball $dx$ \\
        \midrule
        $t$ &  152 & \textcolor{green!60!black}{$+6$} \\
        $t+1$ & 146 & \textcolor{red!60!black}{$-6$}\\
    \end{tabular}
\caption*{\textbf{(c)} Trajectory reveals a bounce at $x=152$, indicating the presence of a wall.}
\end{subfigure}
\end{minipage}

\caption{\textbf{LLM-Guided Code Refinement.} Given a partially specified policy (top left), the LLM optimizer (OptoPrime) uses trajectory data (right) to infer missing environment constants and complete both the docstring and function logic to enable accurate trajectory prediction.}
\label{fig:llm-infered-code}
\end{figure}



\section{Deep RL Results}
\label{sec:app:deep-rl-result}
Atari results vary widely across papers, and many state-of-the-art deep RL models are not open-source. To ensure consistency, the numbers reported in Table \ref{tab:atari-main-table} are from CleanRL \cite{huang2022cleanrl}, the published ICLR blog post \cite{huang202237}, and the public experiment log\footnote{\url{https://wandb.ai/cleanrl/cleanrl.benchmark/reports/Atari--VmlldzoxMTExNTI}}. Runtime is computed from the Weights \& Biases log. For Breakout and Space Invaders, we reported the full training duration; for Pong, the RL policy plateaued before the experiment finished, so we reported the time from the launch of the experiment to peak performance timestep.

Baseline results in Table \ref{tab:atari-main-table} use 8 parallel environments. Faster implementations exist—e.g., Apex-DQN \cite{DBLP:conf/iclr/HorganQBBHHS18} and EnvPool—with 32–64 environments, A2C can solve Breakout in 33 minutes\footnote{See Appendix~\ref{sec:app:deep-rl-result}.}. Our approach uses only a single environment and no specialized speed optimizations.

\section{Atari Agents Code}

Figures \ref{fig:initial_code_pong}, \ref{fig:initial_code_breakout_1}, \ref{fig:initial_code_breakout_2}, \ref{fig:initial_code_breakout_3}, \ref{fig:initial_code_space_invaders_1}, and
\ref{fig:initial_code_space_invaders_2} show the initial code for Pong, Breakout, and Space Invaders. Figures \ref{fig:best_code_pong_1}, \ref{fig:best_code_pong_2}, \ref{fig:best_code_breakout_1}, \ref{fig:best_code_breakout_2}, \ref{fig:best_code_space_invaders_1},
and \ref{fig:best_code_space_invaders_2} show the best learned code for Pong, Breakout, and Space Invaders. 

\begin{figure}[htbp]
\centering
\begin{lstlisting}[language=Python,
breaklines=true,
showstringspaces=false,
basicstyle=\fontsize{8}{9}\selectfont\ttfamily,
numbers=left,
escapechar=!]
@trace.model
class Policy(Module):
    def __call__(self, obs):
        predicted_ball_y = self.predict_ball_trajectory(obs)
        action = self.select_action(predicted_ball_y, obs)
        return action

    @trace.bundle(trainable=True)
    def predict_ball_trajectory(self, obs):
        """
        Predict the y-coordinate where the ball will intersect with the player's paddle by calculating its trajectory,
        using ball's (x, y) and (dx, dy) and accounting for bounces off the top and bottom walls.
    
        Game Setup:
        - Screen dimensions: The game screen has boundaries where the ball bounces
          - Top boundary: y=30
          - Bottom boundary: y=190
        - Paddle positions:
          - Player paddle: right side of screen (x = 140)
          - Enemy paddle: left side of screen (x = 16)
    
        Args:
            obs (dict): Dictionary containing object states for "Player", "Ball", and "Enemy".
                       Each object has position (x,y), size (w,h), and velocity (dx,dy).
    
        Returns:
            float: Predicted y-coordinate where the ball will intersect the player's paddle plane.
                  Returns None if ball position cannot be determined.
    
        """
        if 'Ball' in obs:
            return obs['Ball'].get("y", None)
        return None

    @trace.bundle(trainable=True)
    def select_action(self, predicted_ball_y, obs):
        """
        Select the optimal action to move player paddle by comparing current player position and predicted_ball_y.
        
        IMPORTANT Movement Logic:
        - If the player paddle's y position is GREATER than predicted_ball_y: Move DOWN (action 2)
          (because the paddle needs to move downward to meet the ball)
        - If the player paddle's y position is LESS than predicted_ball_y: Move UP (action 3)
          (because the paddle needs to move upward to meet the ball)
        - If the player paddle is already aligned with predicted_ball_y: NOOP (action 0)
          (to stabilize the paddle when it's in position)
        Ensure stable movement to avoid missing the ball when close by.
    
        Args:
            predicted_ball_y (float): predicted y coordinate of the ball or None
            obs(dict): Dictionary of current game state, mapping keys ("Player", "Ball", "Enemy") to values (dictionary of keys ('x', 'y', 'w', 'h', 'dx', 'dy') to integer values)
        Returns:
            int: 0 for NOOP, 2 for DOWN, 3 for UP
        """
    
        if predicted_ball_y is not None and 'Player' in obs:
            return random.choice([2, 3])
        return 0
\end{lstlisting}
\caption{\textbf{Initial Policy for Pong.}}
\label{fig:initial_code_pong}
\end{figure}

\begin{figure}[htbp]
\centering
\begin{lstlisting}[language=Python,
breaklines=true,
showstringspaces=false,
basicstyle=\fontsize{8}{9}\selectfont\ttfamily,
numbers=left,
escapechar=!]
@trace.model
class Policy(Module):
    def __call__(self, obs):
        predicted_ball_y = self.predict_ball_trajectory(obs)
        action = self.select_action(predicted_ball_y, obs)
        return action

    @trace.bundle(trainable=True)
    def predict_ball_trajectory(self, obs):
        """(same as above)"""
        if "Ball" not in obs:
            return None
    
        ball = obs["Ball"]
        ball_x = float(ball.get("x", 0))
        ball_y = float(ball.get("y", 0))
        ball_dx = float(ball.get("dx", 0))
        ball_dy = float(ball.get("dy", 0))
    
        if ball_dx == 0:
            # Special handling for vertical movement
            if ball_dy > 0:
                # Ball moving down
                return min(190.0, ball_y + 4.0)
            elif ball_dy < 0:
                # Ball moving up
                return max(30.0, ball_y - 4.0)
            return ball_y
    
        # Calculate time to reach paddle
        paddle_x = 140.0
        time_to_paddle = (paddle_x - ball_x) / ball_dx
    
        # Calculate predicted y without bounces
        predicted_y = ball_y + ball_dy * time_to_paddle
    
        # Handle bounces with improved precision
        while predicted_y < 30 or predicted_y > 190:
            if predicted_y < 30:
                predicted_y = 60.0 - predicted_y  # Reflect off top
            elif predicted_y > 190:
                predicted_y = 380.0 - predicted_y  # Reflect off bottom
    
        # Adjust prediction near boundaries
        if predicted_y < 40:
            predicted_y = 40.0
        elif predicted_y > 180:
            predicted_y = 180.0
    
        return predicted_y
\end{lstlisting}
\caption{\textbf{Best Learned Policy for Pong (Part 1).}}
\label{fig:best_code_pong_1}
\end{figure}

\begin{figure}[htbp]
\centering
\begin{lstlisting}[language=Python,
breaklines=true,
showstringspaces=false,
basicstyle=\fontsize{8}{9}\selectfont\ttfamily,
numbers=left,
escapechar=!]

    # continued from above...

    @trace.bundle(trainable=True)
    def select_action(self, predicted_ball_y, obs):
        """(same as above)"""
        if predicted_ball_y is None or "Player" not in obs or "Ball" not in obs:
            return 0
    
        paddle_y = float(obs["Player"].get("y", 0))
        paddle_h = float(obs["Player"].get("h", 15))  # Default paddle height
    
        # Calculate center of paddle with improved precision
        paddle_center = paddle_y + paddle_h / 2.0
    
        ball = obs["Ball"]
        ball_x = float(ball.get("x", 0))
        ball_dx = float(ball.get("dx", 0))
        ball_dy = float(ball.get("dy", 0))
    
        # Base tolerance increased for faster response
        base_tolerance = 4.0
    
        # Distance-based momentum - be more aggressive when ball is close
        distance = abs(140.0 - ball_x)
        distance_factor = max(0.5, min(2.0, distance / 70.0))  # Scale with distance
    
        # Velocity-based momentum
        speed_momentum = min(abs(ball_dy) / 2.0, 3.0)
    
        # Combined adaptive tolerance
        tolerance = base_tolerance * distance_factor + speed_momentum
    
        # Early movement when ball is far and moving slowly
        if distance > 100 and abs(ball_dy) < 2:
            tolerance *= 0.5
    
        # Special handling for straight ball movement
        if ball_dx == 0:
            if abs(ball_dy) > 0:
                # Move towards predicted intersection more aggressively
                tolerance *= 0.5
    
        # Tighter tolerance near paddle edges
        if paddle_y < 40 or paddle_y > 180:
            tolerance *= 0.7
    
        # Decision making with improved positioning
        diff = paddle_center - predicted_ball_y
        if abs(diff) < tolerance:
            return 0  # Stay in position
        elif diff > 0:
            return 2  # Move down
        else:
            return 3  # Move up
\end{lstlisting}
\caption{\textbf{Best Learned Policy for Pong (Part 2).}}
\label{fig:best_code_pong_2}
\end{figure}

\begin{figure}[htbp]
\centering
\begin{lstlisting}[language=Python,
breaklines=true,
showstringspaces=false,
basicstyle=\fontsize{8}{9}\selectfont\ttfamily,
numbers=left,
escapechar=!]
@trace.model
class Policy(Module):
    def __call__(self, obs):
        pre_ball_x = self.predict_ball_trajectory(obs)
        target_paddle_pos = self.generate_paddle_target(pre_ball_x, obs)
        action = self.select_paddle_action(target_paddle_pos, obs)
        return action

    @trace.bundle(trainable=True)
    def generate_paddle_target(self, pre_ball_x, obs):
        """
        Calculate the optimal x coordinate to move the paddle to catch the ball (at predicted_ball_x)
        and deflect the ball to hit bricks with higher scores in the brick wall.

        Logic:
        - Prioritize returning the ball when the ball is coming down (positive dy)
        - The brick wall consists of 6 vertically stacked rows from top to bottom:
          - Row 1 (top): Red bricks (7 pts)
          - Row 2: Orange (7 pts)
          - Row 3: Yellow (4 pts)
          - Row 4: Green (4 pts)
          - Row 5: Aqua (1 pt)
          - Row 6 (bottom): Blue (1 pt)
         - Strategic considerations:
          - Breaking lower bricks can create paths to reach higher-value bricks above
          - Creating vertical tunnels through the brick wall is valuable as it allows
            the ball to reach and bounce between high-scoring bricks at the top
          - Balance between safely returning the ball and creating/utilizing tunnels
            to access high-value bricks
        - Ball speed increases when hitting higher bricks, making it harder to catch

        Args:
            pre_ball_x (float): predicted x coordinate of the ball intersecting with the paddle or None
            obs (dict): Dictionary containing object states for "Player", "Ball", and blocks "{color}B" (color in [R/O/Y/G/A/B]).
                       Each object has position (x,y), size (w,h), and velocity (dx,dy).
        Returns:
            float: Predicted x-coordinate to move the paddle to. 
                Returns None if ball position cannot be determined.
        """
        if pre_ball_x is None or 'Ball' not in obs:
            return None

        return None
\end{lstlisting}
\caption{\textbf{Initial Policy for Breakout (Part 1).}}
\label{fig:initial_code_breakout_1}
\end{figure}

\begin{figure}[htbp]
\centering
\begin{lstlisting}[language=Python,
breaklines=true,
showstringspaces=false,
basicstyle=\fontsize{8}{9}\selectfont\ttfamily,
numbers=left,
escapechar=!]
    # continued from above...

    @trace.bundle(trainable=True)
    def predict_ball_trajectory(self, obs):
        """
        Predict the x-coordinate where the ball will intersect with the player's paddle by calculating its trajectory,
        using ball's (x, y) and (dx, dy) and accounting for bounces off the right and left walls.

        Game setup: 
        - Screen dimensions: The game screen has left and right walls and brick wall where the ball bounces 
          - Left wall: x=9
          - Right wall: x=152
        - Paddle positions:
          - Player paddle: bottom of screen (y=189)
        - Ball speed:
          - Ball deflects from higher-scoring bricks would have a higher speed and is harder to catch.
        - The paddle would deflect the ball at different angles depending on where the ball lands on the paddle
        
        Args:
            obs (dict): Dictionary containing object states for "Player", "Ball", and blocks "{color}B" (color in [R/O/Y/G/A/B]).
                       Each object has position (x,y), size (w,h), and velocity (dx,dy).
        Returns:
            float: Predicted x-coordinate where the ball will intersect the player's paddle plane.
                  Returns None if ball position cannot be determined.
        """
        if 'Ball' not in obs:
            return None
\end{lstlisting}
\caption{\textbf{Initial Policy for Breakout (Part 2).}}
\label{fig:initial_code_breakout_2}
\end{figure}

\begin{figure}[htbp]
\centering
\begin{lstlisting}[language=Python,
breaklines=true,
showstringspaces=false,
basicstyle=\fontsize{8}{9}\selectfont\ttfamily,
numbers=left,
escapechar=!]
    # continued from above...

    @trace.bundle(trainable=True)
    def select_paddle_action(self, target_paddle_pos, obs):
        """
        Select the optimal action to move player paddle by comparing current player position and target_paddle_pos.

        Movement Logic:
        - If the player paddle's center position is GREATER than target_paddle_pos: Move LEFT (action 3)
        - If the player paddle's center position is LESS than target_paddle_pos: Move RIGHT (action 2)
        - If the player paddle is already aligned with target_paddle_pos: NOOP (action 0)
          (to stabilize the paddle when it's in position)
        Ensure stable movement to avoid missing the ball when close by.

        Args:
            target_paddle_pos (float): predicted x coordinate of the position to best position the paddle to catch the ball,
                and hit the ball to break brick wall.
            obs (dict): Dictionary containing object states for "Player", "Ball", and blocks "{color}B" (color in [R/O/Y/G/A/B]).
                Each object has position (x,y), size (w,h), and velocity (dx,dy).
        Returns:
            int: 0 for NOOP, 2 for RIGHT, 3 for LEFT
        """
        if target_paddle_pos is None or 'Player' not in obs:
            return 0
            
        paddle = obs['Player']
        paddle_x = paddle['x']
        paddle_w = paddle['w']
        paddle_center = paddle_x + (paddle_w / 2)
        
        # Add deadzone to avoid oscillation
        deadzone = 2
        if abs(paddle_center - target_paddle_pos) < deadzone:
            return 0  # NOOP if close enough
        elif paddle_center > target_paddle_pos:
            return 3  # LEFT
        else:
            return 2  # RIGHT
\end{lstlisting}
\caption{\textbf{Initial Policy for Breakout (Part 3).}}
\label{fig:initial_code_breakout_3}
\end{figure}

\begin{figure}[htbp]
\centering
\begin{lstlisting}[language=Python,
breaklines=true,
showstringspaces=false,
basicstyle=\fontsize{8}{9}\selectfont\ttfamily,
numbers=left,
escapechar=!]
@trace.model
class Policy(Module):
    def __call__(self, obs):
        pre_ball_x = self.predict_ball_trajectory(obs)
        target_paddle_pos = self.generate_paddle_target(pre_ball_x, obs)
        action = self.select_paddle_action(target_paddle_pos, obs)
        return action

    @trace.bundle(trainable=True)
    def generate_paddle_target(self, pre_ball_x, obs):
        """(same as above)"""
        if pre_ball_x is None or "Ball" not in obs:
            return None
    
        paddle = obs["Player"]
        paddle_w = paddle["w"]
        ball = obs["Ball"]
        ball_dx = ball["dx"]
        ball_y = ball["y"]
    
        # Find gaps in brick rows to aim for
        gaps = []
        for y in [87, 81, 75, 69, 63, 57]:  # Bottom to top rows
            row_blocks = [
                b
                for b in obs.get(
                    f'{"B" if y == 87 else "A" if y == 81 else "G" if y == 75 else "Y" if y == 69 else "O" if y == 63 else "R"}B',
                    [],
                )
            ]
            if not row_blocks:
                continue
            for i in range(len(row_blocks)):
                if i > 0:
                    gap_start = row_blocks[i - 1]["x"] + row_blocks[i - 1]["w"]
                    gap_end = row_blocks[i]["x"]
                    if gap_end - gap_start > 6:  # Min gap width
                        gaps.append((gap_start + gap_end) / 2)
    
        # Base offset that ensures reliable ball return
        base_offset = -3 if ball_dx > 0 else 3
    
        # Adjust offset based on ball height and gaps
        if ball_y < 90:  # Ball near brick wall
            if gaps:  # Aim for closest gap
                closest_gap = min(gaps, key=lambda x: abs(x - pre_ball_x))
                if abs(closest_gap - pre_ball_x) < 30:  # Gap within reach
                    return closest_gap
    
        # When ball is low or no good gaps available, focus on safe return
        return pre_ball_x + base_offset
\end{lstlisting}
\caption{\textbf{Best Learned Policy for Breakout (Part 1).}}
\label{fig:best_code_breakout_1}
\end{figure}

\begin{figure}[htbp]
\centering
\begin{lstlisting}[language=Python,
breaklines=true,
showstringspaces=false,
basicstyle=\fontsize{8}{9}\selectfont\ttfamily,
numbers=left,
escapechar=!]
    # continued from above...

    @trace.bundle(trainable=True)
    def predict_ball_trajectory(self, obs):
        """(same as above)"""
        if "Ball" not in obs:
            return None

        ball = obs["Ball"]
        ball_x = ball["x"]
        ball_y = ball["y"]
        ball_dx = ball["dx"]
        ball_dy = ball["dy"]
    
        # If ball is not moving or moving up, can't predict
        if ball_dy <= 0:
            return None
    
        # Calculate time to reach paddle
        paddle_y = 189
        time_to_paddle = (paddle_y - ball_y) / ball_dy
    
        # Calculate x position considering wall bounces
        num_bounces = 0
        pred_x = ball_x + (ball_dx * time_to_paddle)
    
        while pred_x < 9 or pred_x > 152:
            if pred_x < 9:
                pred_x = 9 + (9 - pred_x)
                num_bounces += 1
            elif pred_x > 152:
                pred_x = 152 - (pred_x - 152)
                num_bounces += 1
            if num_bounces > 10:  # Avoid infinite bounces
                return None
    
        return pred_x
        
    @trace.bundle(trainable=True)
    def select_paddle_action(self, target_paddle_pos, obs):
        """(same as above)"""
        if target_paddle_pos is None or "Player" not in obs:
            return 0
    
        paddle = obs["Player"]
        paddle_x = paddle["x"]
        paddle_w = paddle["w"]
        paddle_center = paddle_x + (paddle_w / 2)
        ball = obs.get("Ball", {})
    
        # Adaptive deadzone based on ball position and speed
        base_deadzone = 3
        ball_y = ball.get("y", 189)
        ball_dy = abs(ball.get("dy", 0))
    
        # Larger deadzone for faster balls and higher positions
        height_factor = (189 - ball_y) / 189
        speed_factor = ball_dy / 4
        deadzone = base_deadzone * (1 + height_factor + speed_factor)
    
        if abs(paddle_center - target_paddle_pos) < deadzone:
            return 0  # NOOP if close enough
        elif paddle_center > target_paddle_pos:
            return 3  # LEFT
        else:
            return 2  # RIGHT
\end{lstlisting}
\caption{\textbf{Best Learned Policy for Breakout (Part 2).}}
\label{fig:best_code_breakout_2}
\end{figure}

\begin{figure}[htbp]
\centering
\begin{lstlisting}[language=Python,
breaklines=true,
showstringspaces=false,
basicstyle=\fontsize{8}{9}\selectfont\ttfamily,
numbers=left,
escapechar=!]
@trace.model
class Policy(Module):
    def __call__(self, obs):
        shoot_decision = self.decide_shoot(obs)
        move_decision = self.decide_movement(obs)
        return self.combine_actions(shoot_decision, move_decision)

    @trace.bundle(trainable=True)
    def combine_actions(self, shoot, movement):
        '''
        Combine shooting and movement decisions into final action.
        
        Args:
            shoot (bool): Whether to shoot
            movement (int): Movement direction
        
        Action mapping:
        - 0: NOOP (no operation)
        - 1: FIRE (shoot without moving)
        - 2: RIGHT (move right without shooting)
        - 3: LEFT (move left without shooting)
        - 4: RIGHT+FIRE (move right while shooting)
        - 5: LEFT+FIRE (move left while shooting)
        
        Returns:
            int: Final action (0: NOOP, 1: FIRE, 2: RIGHT, 3: LEFT, 4: RIGHT+FIRE, 5: LEFT+FIRE)
        '''
        
        if shoot and movement > 0:
            return 4  # RIGHT+FIRE
        elif shoot and movement < 0:
            return 5  # LEFT+FIRE
        elif shoot:
            return 1  # FIRE
        elif movement > 0:
            return 2  # RIGHT
        elif movement < 0:
            return 3  # LEFT
        return 0  # NOOP
\end{lstlisting}
\caption{\textbf{Initial Policy for Space Invaders (Part 1).}}
\label{fig:initial_code_space_invaders_1}
\end{figure}

\begin{figure}[htbp]
\centering
\begin{lstlisting}[language=Python,
breaklines=true,
showstringspaces=false,
basicstyle=\fontsize{8}{9}\selectfont\ttfamily,
numbers=left,
escapechar=!]
    # continued from above...

    @trace.bundle(trainable=True)
    def decide_movement(self, obs):
        '''
        Decide movement direction based on enemy positions and projectiles.
         
        Args:
            obs (dict): Game state observation containing object states for "Player", "Shield0", "Shield1", "Alien0", "Alien1", etc.
            Each object has position (x,y), size (w,h), and velocity (dx,dy).
            Player bullets have negative dy velocity and alien bullets have positive dy velocity
        
        Strategy tips:
        - Move to dodge enemy projectiles
        - Position yourself under aliens to shoot them
        - Stay away from the edges of the screen
        - Consider moving toward areas with more aliens to increase score
        
        Returns:
            int: -1 for left, 1 for right, 0 for no movement
        '''

        player = obs['Player']

        return random.choice([-1, 0, 1])

    @trace.bundle(trainable=True)
    def decide_shoot(self, obs):
        '''
        Decide whether to shoot based on enemy positions and existing projectiles.
         
        Args:
            obs (dict): Game state observation containing object states for "Player", "Shield0", "Shield1", "Alien0", "Alien1", etc.
            Each object has position (x,y), size (w,h), and velocity (dx,dy).
            Player bullets have negative dy velocity and alien bullets have positive dy velocity
        
        Strategy tips:
        - You can only have one missile at a time
        - Try to shoot when aliens are aligned with your ship
        - Prioritize shooting at lower aliens as they're closer to you
        - Consider the movement of aliens when deciding to shoot
        
        Returns:
            bool: True if should shoot, False otherwise
        '''
        
        # There can only be one player bullet on the field at a time
        # Check for player bullets (which have negative dy velocity)
        for key, obj in obs.items():
            if key.startswith('Bullet') and obj.get('dy', 0) < 0:
                return False
            
        player = obs['Player']
        for key, obj in obs.items():
            if key.startswith('Alien'):
                # Check if alien is aligned with player (within 5 pixels)
                if abs(obj['x'] - player['x']) < 5:
                    # Prioritize lower aliens (higher y value)
                    if obj['y'] > 60:  # Adjust this threshold as needed
                        return True
        return False
\end{lstlisting}
\caption{\textbf{Initial Policy for Space Invaders (Part 2).}}
\label{fig:initial_code_space_invaders_2}
\end{figure}

\begin{figure}[htbp]
\centering
\begin{lstlisting}[language=Python,
breaklines=true,
showstringspaces=false,
basicstyle=\fontsize{8}{9}\selectfont\ttfamily,
numbers=left,
escapechar=!]
@trace.model
class Policy(Module):
    def __call__(self, obs):
        shoot_decision = self.decide_shoot(obs)
        move_decision = self.decide_movement(obs)
        return self.combine_actions(shoot_decision, move_decision)

    @trace.bundle(trainable=True)
    def combine_actions(self, shoot, movement):
        # same as above

    @trace.bundle(trainable=True)
    def decide_movement(self, obs):
        """(same as above)"""
    
        import random
    
        player = obs["Player"]
        move = 0
        threat_left = 0
        threat_right = 0
        aliens_left = 0
        aliens_right = 0
        screen_width = 160  # Assuming standard Space Invaders screen width
    
        for key, obj in obs.items():
            if key.startswith("Alien"):
                if obj["x"] < player["x"]:
                    aliens_left += 1
                else:
                    aliens_right += 1
            elif key.startswith("Bullet") and obj["dy"] > 0:  # Enemy bullet
                if obj["x"] < player["x"]:
                    threat_left += 1
                else:
                    threat_right += 1
                # Consider vertical position of bullets
                if abs(obj["x"] - player["x"]) < 10 and obj["y"] > player["y"] - 30:
                    move = 1 if obj["x"] < player["x"] else -1
    
        # Move away from threats if no immediate vertical threat
        if move == 0:
            if threat_left > threat_right:
                move = 1
            elif threat_right > threat_left:
                move = -1
            # If no immediate threat, move towards more aliens
            elif aliens_left > aliens_right:
                move = -1
            elif aliens_right > aliens_left:
                move = 1
    
        # Stay away from screen edges
        if player["x"] < 10 and move == -1:
            move = 1
        elif player["x"] > screen_width - 10 and move == 1:
            move = -1
    
        # Add small random movement
        if random.random() < 0.1:
            move = random.choice([-1, 0, 1])
    
        return move
    
\end{lstlisting}
\caption{\textbf{Best Learned Policy for Space Invaders (Part 1).}}
\label{fig:best_code_space_invaders_1}
\end{figure}

\begin{figure}[htbp]
\centering
\begin{lstlisting}[language=Python,
breaklines=true,
showstringspaces=false,
basicstyle=\fontsize{8}{9}\selectfont\ttfamily,
numbers=left,
escapechar=!]
    # continued from above...

    @trace.bundle(trainable=True)
    def decide_shoot(self, obs):
        """(same as above)"""
    
        # There can only be one player bullet on the field at a time
        # Check for player bullets (which have negative dy velocity)
        for key, obj in obs.items():
            if key.startswith("Bullet") and obj.get("dy", 0) < 0:
                return False
    
        player = obs["Player"]
        closest_alien_distance = float("inf")
        closest_alien = None
    
        for key, obj in obs.items():
            if key.startswith("Alien"):
                distance = abs(obj["x"] - player["x"])
                if distance < closest_alien_distance:
                    closest_alien_distance = distance
                    closest_alien = obj
    
        if closest_alien:
            # Check if alien is aligned with player (within 15 pixels)
            if abs(closest_alien["x"] - player["x"]) < 15:
                # Prioritize lower aliens (higher y value)
                if (
                    closest_alien["y"] > 30
                ):  # Lowered threshold for more aggressive shooting
                    return True
        return False
\end{lstlisting}
\caption{\textbf{Best Learned Policy for Space Invaders (Part 2).}}
\label{fig:best_code_space_invaders_2}
\end{figure}


\end{document}